\documentclass[runningheads]{llncs}

\usepackage[final,year=2024,ID=3230]{eccv}
\usepackage{eccvabbrv}

\usepackage{url}
\usepackage{ftnxtra}
\usepackage{fnpos}
\usepackage{pifont}

\usepackage{graphicx}
\usepackage{animate}
\usepackage{booktabs}
\usepackage{fancyhdr}

\usepackage{float}
\usepackage[accsupp]{axessibility}

\usepackage{xspace}
\makeatletter
\DeclareRobustCommand\onedot{\futurelet\@let@token\@onedot}
\def\@onedot{\ifx\@let@token.\else.\null\fi\xspace}

\makeatother

\usepackage{xcolor}
\usepackage{multirow}
\usepackage{enumitem}
\usepackage{amsmath}

\usepackage[pagebackref,breaklinks,colorlinks,citecolor=eccvblue]{hyperref}
\usepackage{orcidlink}

\newcommand{\cococo}{{\textbf{CoCoCo}}}
\newcommand{\CoCoCo}{{\textbf{CoCoCo} }}

\definecolor{blue}{RGB}{63,197,242}

\begin{document}

\title{CoCoCo: Improving Text-Guided Video Inpainting for Better \textcolor{blue}{Co}nsistency, \textcolor{blue}{Co}ntrollability and \textcolor{blue}{Co}mpatibility}

\titlerunning{CoCoCo}

\author{Bojia Zi\inst{1} \and
Shihao Zhao\inst{3} \and
Xianbiao Qi\thanks{is the corresponding author.}\inst{2} \and
Jianan Wang\inst{2} \and
Yukai Shi\inst{4} \and 
Qianyu Chen\inst{1} \and
Bin Liang\inst{1} \and 
Kam-Fai Wong\inst{1} \and 
Lei Zhang\inst{2} }

\authorrunning{B.~Zi et al.}

\institute{The Chinese University of Hong Kong \\
\email{\{bjzi,qychen,kfwong\}@se.cuhk.edu.hk} 
\email{bin.liang@cuhk.edu.hk} \and
International Digital Economy Academy (IDEA) \\
\email{\{qixianbiao,wangjianan,leizhang\}@idea.edu.cn} \and
The University of Hong Kong\\
\email{shzhao@cs.hku.hk} \and
Tsinghua University \\
\email{shiyk22@mails.tsinghua.edu.cn}}

\maketitle

\begin{figure}[h]
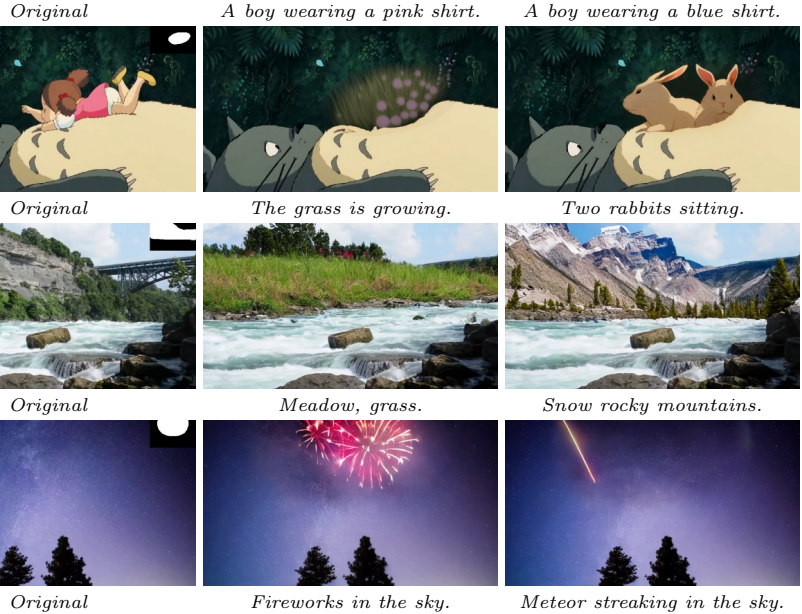

  \centering
\begin{subfigure}{0.32\textwidth}
      \animategraphics[width=1\textwidth]{9}{data/gibuli/gibuli_org/image_}{0}{13}
      \subcaption*{\emph{\scriptsize Original}}
       
\end{subfigure}
\begin{subfigure}{0.32\textwidth}
      \animategraphics[width=1\textwidth]{9}{data/gibuli/video_pink_image}{0}{13}
      \subcaption*{\emph{\scriptsize A boy wearing a pink shirt.}}
\end{subfigure}
\begin{subfigure}{0.32\textwidth}
      \animategraphics[width=1\textwidth]{9}{data/gibuli/gibuli/video_}{0}{13}
      \subcaption*{\emph{\scriptsize A boy wearing a blue shirt.}}
\end{subfigure}
\begin{subfigure}{0.32\textwidth}
      \animategraphics[width=1\textwidth]{6}{data/longcat/image_}{0}{7}
      \subcaption*{\emph{\scriptsize Original}}
\end{subfigure}
\begin{subfigure}{0.32\textwidth}
      \animategraphics[width=1\textwidth]{6}{data/longcat/video_grass_ghibli_image_}{0}{7}
      \subcaption*{\emph{\scriptsize The grass is growing.}}
\end{subfigure}
\begin{subfigure}{0.32\textwidth}
      \animategraphics[width=1\textwidth]{6}{data/longcat/video_rabbit_ghibli_image_}{0}{7}
      \subcaption*{\emph{\scriptsize Two rabbits sitting.}}
\end{subfigure}
\begin{subfigure}{0.32\textwidth}
      \animategraphics[width=1\textwidth]{8}{data/video_1/image_}{0}{11}
      \subcaption*{\emph{\scriptsize Original}}
\end{subfigure}
\begin{subfigure}{0.32\textwidth}
      \animategraphics[width=1\textwidth]{8}{data/meadow2/video_meadow_image_}{0}{11}
      \subcaption*{\emph{\scriptsize Meadow, grass.}}
\end{subfigure}
\begin{subfigure}{0.32\textwidth}
      \animategraphics[width=1\textwidth]{8}{data/snow_mountain/video_snow_mountain_image_}{0}{11}
      \subcaption*{\emph{\scriptsize Snow rocky mountains.}}
\end{subfigure}
\begin{subfigure}{0.32\textwidth}
      \animategraphics[width=1\textwidth]{10}{data/video_2/image_}{0}{15}
      \subcaption*{\emph{\scriptsize Original}}
\end{subfigure}
\begin{subfigure}{0.32\textwidth}
      \animategraphics[width=1\textwidth]{10}{data/fireworks/video_fireworks_image_}{0}{15}
      \subcaption*{\emph{\scriptsize Fireworks in the sky.}}
\end{subfigure}
\begin{subfigure}{0.32\textwidth}
      \animategraphics[width=1\textwidth]{10}
    {data/metoer/video_metoer_image_}{0}{15}
      \subcaption*{\emph{\scriptsize Meteor streaking in the sky.}}
\end{subfigure}
\caption{The inpainting results of our \CoCoCo method. The first and second rows are the results of our model with CounterfeitV30 T2I personalized model plugged in, and the last two rows are the results only with our model. \emph{Best viewed with Acrobat Reader. Click the images to play the animation clips.}}
\label{fig:gif_example1}
\end{figure}
\begin{abstract}
Recent advancements in video generation have been remarkable, yet many existing methods struggle with issues of consistency and poor text-video alignment. Moreover, the field lacks effective techniques for text-guided video inpainting, a stark contrast to the well-explored domain of text-guided image inpainting.
To this end, this paper proposes a novel text-guided video inpainting model that achieves better consistency, controllability
and compatibility. Specifically, we introduce a simple but efficient motion capture module to preserve motion consistency, and design an instance-aware region selection instead of a random region selection to obtain better textual controllability, and utilize a novel strategy to inject some personalized models into our \CoCoCo model and thus obtain better model compatibility. Extensive experiments show that our model can generate high-quality video clips. Meanwhile, our model shows better motion consistency, textual controllability and model compatibility. More details are shown in \href{https://cococozibojia.github.io}{cococozibojia.github.io}.

\end{abstract}
\section{Introduction}

The field of video generation~\cite{sora_videoworldsimulators2024,
imagenvideo_ho2022imagen, song2022diffusion, gen1_esser2023structure, gen2, stablediffusion_blattmann2023align, videocrafter2_chen2024videocrafter2, videopoet_kondratyuk2023videopoet, wang2023videocomposer, cogvideo_hong2022cogvideo, pikalab} has recently garnered significant attention from the public. It enables the creation of video content through the use of text prompts. SORA~\cite{sora_videoworldsimulators2024}, Gen2~\cite{gen2}, VideoPoet~\cite{videopoet_kondratyuk2023videopoet}, and Pika~\cite{pikalab} have propelled video generation capability to a new level, allowing for the creation of high-resolution videos of long duration with high visual effects occasionally. Despite the success of these closed-source methods, current open-source video generation still faces a lot of complaints from the users, including issues of low consistency, poor textual controllability, and poor visual quality. These disadvantages still limits the potential applications. 

Text-guided video inpainting is an effective way to modify undesired content in video. 
Different from the text-guided image inpainting that generates visual content within masked regions based on text prompts, text-guided video inpainting produces dynamic video content across frames, guided by prompts. Although the current text-guided video inpainting method, notably AVID~\cite{avid_zhang2023avid}, has demonstrated impressive results, it faces several challenges. 1). It directly utilizes and fine-tunes the motion block from AnimateDiff~\cite{animatediff_guo2023animatediff} \textit{without adequately addressing text-video alignment and motion consistency across frames}. 2). its training approach, which involves the use of a random mask selection, will lead to \textit{a mismatch between the masked region and the specified text prompt}, especially when the object described by the prompt might only occupy a small portion of the frame. 3). it does not show good ability to integrate with personalized Text-to-Image (T2I) models.

To mitigate the aforementioned three drawbacks, we introduce \cococo, a novel method to improve text-guided video inpaintng  for better \textit{consistency, controllability,
and compatibility}. Our improvements lie in three aspects. First, we improve the motion capture module  by adding two attention layers: a damped global attention layer and a textual cross-attention layer. Second, we design a new mask region selection strategy, termed instance-aware region selection. Specifically, we use Grounding DINO~\cite{groundingdino_liu2023grounding} to detect the given prompt for the first frame in the training clip. Subsequently, we use the tokenspan to keep the consistency of detected phrases corresponding to an object in the rest of frames. 
In the training stage, we randomly sample from three types of data with different probabilities, including clip with precise masks, clip with random masks and clip with null prompts. Besides, 
we propose a strategy to transform an image generation model to make it compatible with our video inpainting model inspired by the thought of the task vector~\cite{taskvector_ilharco2022editing}. In this way, we can combine some personalized T2I models with our model, and thus create the customized content in the masked region of the given video.

Our contributions in this paper can be summarised into three  aspects:
\begin{itemize}[leftmargin=*]
    \item We propose a novel motion capture module. It consists of three types attention block, including two previously used temporal attention layers, a newly introduced damped global attention layer and a textual cross attention layer. The new motion capture module can enable the model to have better motion consistency and text-video controllability.
    \item We design a new instance-aware region selection strategy instead of a random mask selection strategy used in previous method~\cite{avid_zhang2023avid}. The new strategy can help the model achieve better text-video controllability. 
    \item We introduce a novel strategy to transform some personalized generation models, and then plug them into our our text-guided video inpainting model. This strategy can enhance the compatibility of our model.
\end{itemize}

\section{Related Work}

\textbf{Image Generation.} Image generation has received a huge amount of attention from the public\cite{dalle1_ramesh2021zero, glide_nichol2021glide, dalle2_ramesh2022hierarchical, dalle3_betker2023improving, imagenvideo_ho2022imagen, stablediffusion_blattmann2023align, ddpm_ho2020denoising, ddim_song2020denoising, sdeforgm_song2020score} in recent years. DALL-E~\cite{dalle1_ramesh2021zero} uses a VQ-VAE~\cite{vqvae_van2017neural} to encode the image into some discrete words, and thus can use GPT~\cite{gpt1_radford2018improving} model to train a generation model. GLIDE~\cite{glide_nichol2021glide} introduces Classifier-Free Guidance to improve text-alignment ability and promote better generation results. 
DALL-E2 utilizes the CLIP~\cite{clip_radford2021learning} feature to improve the text-image alignment. Imagen~\cite{imagen_saharia2022photorealistic} uses a cascade architecture to generate image,  while scaling the language model to billions of parameters. Latent Diffusion Model (LDM)~\cite{stablediffusion_blattmann2023align} uses a VAE~\cite{vae_kingma2013auto} to encode the input into the continuous latent space and reconstruct the latent codes into the image, and then conduct the diffusion process in latent space instead of the original pixel space. 
Among these generation methods, personalized image generation technique, such as DreamBooth~\cite{dreambooth_ruiz2022dreambooth}, is attractive because it helps people customize an image generation, while only requires low GPU resources and a small amount of private data.

\noindent \textbf{Text-Guided Image Inpainting.} 
There are several works in text-guided image inpainting\cite{paintbyword_andonian2021paint, avrahami2022blended, cogview2_ding2022cogview2, Diffedit_couairon2022diffedit, imageneditor_wang2023imagen}. Paint-By-Word~\cite{paintbyword_andonian2021paint} optimizes to trade off the image consistency and text-alignment. Similarly, Blended Diffusion~\cite{avrahami2022blendedldm} executes CLIP-guided diffusion processes concurrently on both foreground and background, subsequently merging the outcomes through element-wise aggregation. CogView2\cite{cogview2_ding2022cogview2} introduces an auto-regressive method for text-guided image inpainting to enhance textual alignment. Additionally, DiffEdit\cite{Diffedit_couairon2022diffedit} innovates with a ``masked yet mask-free'' approach, simultaneously executing masking segmentation and masked diffusion to achieve seamless masked inpainting. Imagen Editor\cite{imageneditor_wang2023imagen} uses a cascaded diffusion model to perform image inpainting by fine-tuning a Imagen model. 

\noindent \textbf{Video Generation.} 
Recently, many video generation methods have been proposed~\cite{imagenvideo_ho2022imagen, gen1_esser2023structure, gen2, pikalab, sora_videoworldsimulators2024, stablevideodiffusion_blattmann2023stable, videopoet_kondratyuk2023videopoet, videocrafter2_chen2024videocrafter2, Dynamicrafter_xing2023dynamicrafter, 
tuneavideo_wu2023tune, khachatryan2023text2video, animatediff_guo2023animatediff,chen2023controlavideo, yin2023dragnuwa, chen2023controlavideo,modelscope_wang2023modelscope, fan2023hierarchical, zhang2023show1, guo2023sparsectrl, jiang2023videobooth, ma2024magicme,
xie2023dreaminpainter,
ma2023follow}. For the closed-source products, such as SORA~\cite{sora_videoworldsimulators2024}, Pika~\cite{pikalab}, VideoPoet~\cite{videopoet_kondratyuk2023videopoet}, Gen1~\cite{gen1_esser2023structure}, and Gen2~\cite{gen2}, they provide impressive visual results with high resolution and long duration. However, the details of their methods are unknown and data for training is unavailable for the public. For the open-source methods, we have also witnessed a rapid development. Tune-A-Video~\cite{tuneavideo_wu2023tune} only adapts a small proportion of parameters and makes slight architecture modifications on the image diffusion models, and can achieve zero-shot video generation. Text2Video-Zero~\cite{khachatryan2023text2video} is a training free method to create videos by editing the latent codes with a predefined affine matrix. AnimateDiff~\cite{animatediff_guo2023animatediff} trains the motion module and keeps image module frozen to adopt the personalized models and can generate videos using  different personalized T2Is models. VideoCrafter~\cite{VideoCrafter1_chen2023videocrafter1, videocrafter2_chen2024videocrafter2} gives way to create high-quality video condition on a prompt or an image. DynamiCrafter~\cite{Dynamicrafter_xing2023dynamicrafter} generates video by incorporating the image into the generative process as a guidance. Stable Video Diffusion~\cite{stablevideodiffusion_blattmann2023stable} utilizes a well-curated pretraining dataset for high-quality video generation by a reasonable captioning and filtering strategies. ModelScope~\cite{modelscope_wang2023modelscope} introduces a text-to-video synthesis model, which incorporates spatial-temporal blocks to keep consistency and thus has smooth movement transitions.

\noindent \textbf{Text-Guided Video Inpainting.} 
Recently, a text-guided video inpainting, termed as AVID~\cite{avid_zhang2023avid}, is proposed. They follows a similar architecture as AnimateDfiff~\cite{animatediff_guo2023animatediff} by initializing the image module with image inpainting models and finetuning motion module initialized with a pretrained AnimateDiff motion module. To get higher visual quality, it injects the textual information into the down blocks and middle block of UNet~\cite{unet_ronneberger2015u} similar to ControlNet\cite{controlnet_zhang2023adding}.

\noindent \textbf{Remark.} \CoCoCo targets a better text-guided video inpainting by improving its motion \textit{consistency} by introducing a new motion capture module, \textit{controllability} via designing an instance-aware region selection strategy instead of random region selection and adding a textual cross-attention  block, and \textit{compatibility} through enabling to plug some after-transformed personalized model.

\begin{figure*}[t]
  \centering
  \includegraphics[width=0.820\textwidth]{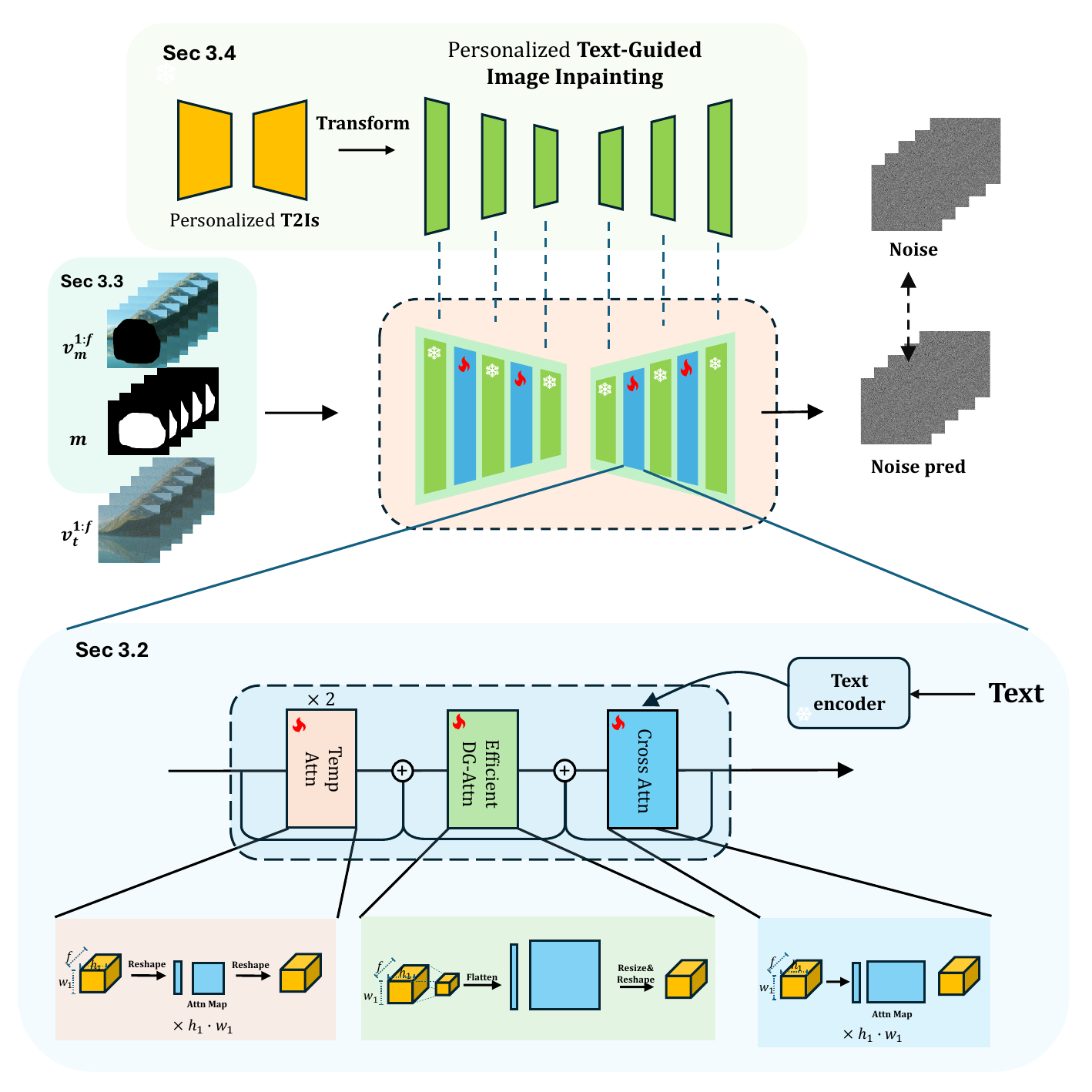} 
  \caption{The overall framework of \cococo. As shown in the figure, \cococo \  has three inputs including masked video, mask, and noised video. As shown in the above of the figure, our model can adapt the text-to-image (T2I) personalized models without model-specific tuning to perform text-guided video inpainting.  The personalized models can be downloaded from the opensource platforms, such as \emph{CivitAI} and \emph{Huggingface}. Meanwhile, as shown in the below of the figure, our model uses a newly introduced motion capture module that consists of three types of attention blocks.}
  \label{fig:framework4}
\end{figure*}

\section{Methodology}
\subsection{The Overall Framework of CoCoCo}
Figure~\ref{fig:framework4} illustrates the overall framework of \cococo. Our framework is built on a UNet~\cite{unet_ronneberger2015u} architecture. In our model, we incorporate a specialized set of layers that have been tuned to better adapt to the nuances of video data. 

As shown in Figure~\ref{fig:framework4}, the input to the model consists of three components: an inpainting mask $m^{1:f}$, the masked video clip $v^{1:f}\odot m^{1:f}$, the noised video clip $\sqrt{\Bar{\alpha_t}} v^{1:f} + \sqrt{1-\Bar{\alpha_t}} \epsilon^{1:f}$. We use VAE~\cite{vae_kingma2013auto} to encode the masked video clip and the noised video clip into latent vectors, and the perform diffusion process on the latent vector, it can largely speedup the training process. 
Similar with the previous work~\cite{animatediff_guo2023animatediff, avid_zhang2023avid}, the frozen block in the model is derived from the image inpainting block. We insert some trainable motion capture modules among frozen modules. As shown in the below of the Figure~\ref{fig:framework4}, the motion capture module is comprised of three distinct types of attention blocks: two temporal attention block, a damped global attention block to preserve motion consistency, and a textual cross-attention block to improve text-video alignment. As shown in the above of the Figure~\ref{fig:framework4}, to achieve compatibility with personalized text2image (T2I) model, we transform the model to make it compatible with our video inpainting model.

\CoCoCo achieves the following three substantial advantages:
\begin{itemize}[leftmargin=*]
    \item \textbf{Consistency in Motion.} We obtain better motion consistency via introduction a damped global attention (DGA) instead of a temporal-only attention used in AVID~\cite{avid_zhang2023avid}. DGA can enable a better global information capture, and thus obtain a better consistency. See Section~\ref{sec:motion_capture}.
    
    \item \textbf{Controllability.} We obtain a better controllability by two ways. First, we introduce an instance-aware region selection strategy to align region and text.
    Second, we increase a textual cross-attention module to distill textual information to enable better textual controllability. See Section~\ref{sec:motion_capture} and ~\ref{sec:instanceawareregionselection}.

    \item \textbf{Compatibility.} We introduce a simple but effective strategy to transform personalized text2image models to that is compatible with our model. See Section~\ref{sec:taskvectorcombination}.
\end{itemize}

\subsection{Motion Capture Module}
\label{sec:motion_capture}
\begin{figure*}[h]
  \centering
  \includegraphics[width=0.75\textwidth]{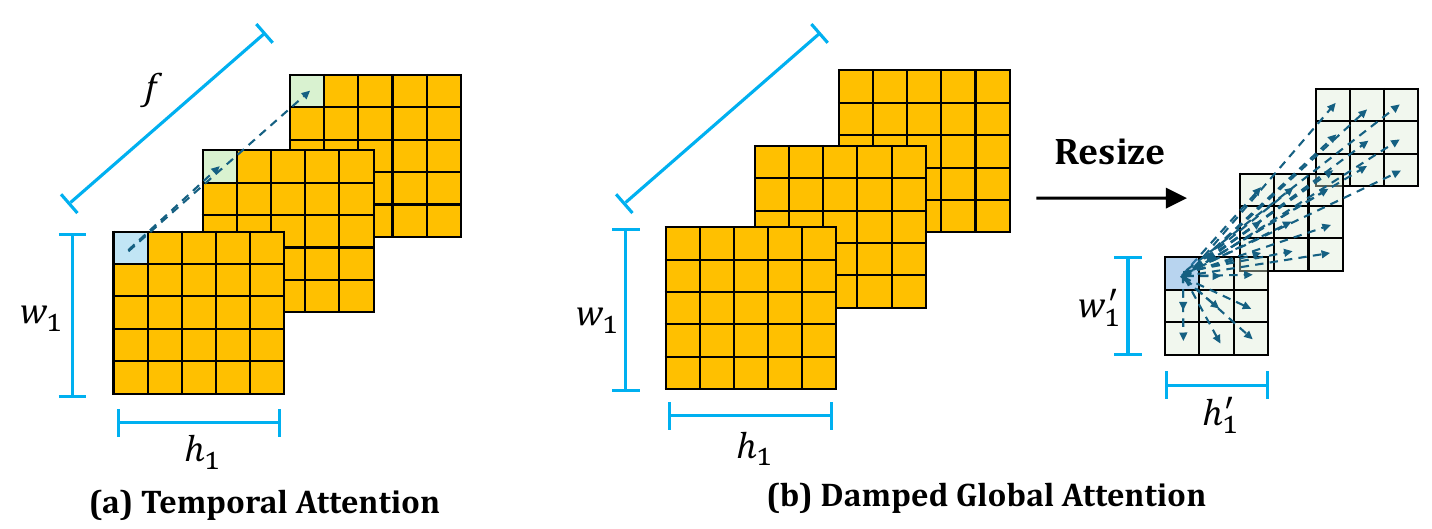} 
  \caption{The comparison between temporal attention and damped global attention. The dotted line indicates the positions that can be attended.}
  \label{fig:motion_modules}
\end{figure*}

In previous works~\cite{animatediff_guo2023animatediff, avid_zhang2023avid}, only two temporal attention mechanisms are employed within their motion blocks to capture motion information, while the spatial block remains unchanged. Although this design  learns motion dynamics, it also brings in some issues.
Firstly, the temporal attention mechanism, as shown in the left of Figure~\ref{fig:motion_modules},  is limited by its inability in attending spatial regions. 
Therefore, the attention mechanism lacks the capacity to grasp global information, resulting in an inability to adapt the inpainting region based on surrounding areas.
Secondly, the motion blocks are not designed to incorporate text guidance, thus overlooking crucial textual cues during motion generation.

To resolve these problems, we propose to insert two attentions after the two temporal attentions, including an efficient damped global attention to capture global motion information, and a textual cross attention to learn the motion under the guidance of text prompts. 

\noindent \textbf{Damped Global Attention.} In this paper, we introduce a simple but effective \textit{damped global attention (DGA)} to improve motion consistency. As shown in Figure~\ref{fig:motion_modules}, in DGA,
we first adjust the spatial dimensions of the input features via resizing the feature mappings from $w_1 \times h_1$ to $w_1' \times h_1'$. Subsequently, we flatten the tensor $z \in \mathbb{R}^{f\times w_1' \times h_1'}$ into a vector $x$, which has a sequence length of $L = f \cdot w_1' \cdot h_1'$. This vector $x$ is then fed into a multi-head  self-attention layer~\cite{transformer_vaswani2017attention}. After processing, the vector is reshaped and resized back to its original dimensions of $f \times w_1 \times h_1$. This strategy can reduce memory consumption, while still grasp global information effectively. 

We illustrate the visual comparison between temporal attention and damped global attention in Figure~\ref{fig:motion_modules}. We further present the comparison between the computing process of two different attention mechanisms as below:

\noindent {Temporal Attention}.
\[
\begin{aligned}
 x \in \mathbb{R}^{b \cdot f \cdot c \cdot w_1 \cdot h_1} \xrightarrow{\text{packed}}  \mathbb{R}^{(b \cdot w_1 \cdot h_1) \cdot f \cdot c} 
\xrightarrow{\text{SA}} \mathbb{R}^{(b \cdot w_1 \cdot h_1) \cdot f \cdot c} 
\xrightarrow{\text{unpacked}} \mathbb{R}^{b \cdot f \cdot c \cdot w_1 \cdot h_1 }
\end{aligned}
\]
\noindent {Our DGA attention}.
\[
\begin{aligned}
x \in \mathbb{R}^{b \cdot f \cdot c \cdot w_1 \cdot h_1} \xrightarrow{\substack{\text{resize } \\ \text{ \& reshape}}}  \mathbb{R}^{b \cdot  (w_1'  \cdot h_1' \cdot f) \cdot c} 
\xrightarrow{\text{SA}}  \mathbb{R}^{b  \cdot (w_1' \cdot h_1' \cdot f) \cdot c}  
\xrightarrow{\substack{\text{resize } \\ \text{ \& reshape}}}
  \mathbb{R}^{b \cdot f \cdot c \cdot w_1 \cdot h_1}
\end{aligned}
\]

\noindent \textbf{Adding Textual Cross Attention for Better Controllability.} In the development of AnimateDiff and AVID, text embeddings are used in spatial layers without fine-tuning any projection parameters. This approach presents a significant limitation: the inability to incorporate motion details from the text prompts. To address this, we've incorporated a textual cross-attention mechanism into our motion capture module, enhancing the representation of motion information. To reduce memory usage, we employ a flattened vector from the visual input \(z^{f\times w_1^{'} \times h_1^{'}}\) as the query, with text embeddings serving as both key and value. This method significantly reduces the attention map's size from \((f \times w_1 \times h_1 \times l_{text})^{2}\) to \((f \times l_{text})^{2}\), where \(l_{text}\) denotes the text embedding length.

For weight initialization of our motion capture module, we 
initialize the temporal attentions with AnimateDiff-v2, while we initialize the remaining DGA and textural cross attention with Kaiming initialization. Regardless of their initialization method, all four attentions are optimized using the same learning rate. \textit{Damped global attention and explicitly adding textual cross-attention can improve motion consistency and text controllability.} 

\subsection{Instance-aware Region Selection for Video Inpainting}
\label{sec:instanceawareregionselection}
A random mask selection is used in the training stage for inpainting in AVID~\cite{avid_zhang2023avid}.  Nevertheless, this approach owns multiple disadvantages. Primarily, videos typically comprise multiple scenes, and directly sampling N frames from a video poses the risk of creating a training clip that includes frames from two distinct scenes. 
Subsequently, training a video diffusion model with such inconsistent clips can lead to instability in training and diminished inpainting consistency. Moreover, employing a random mask selection might fail to cover any object or only partially cover an object, thus cannot learn the motion information relative to the given prompt.

We should promise the mask consistency among frames. To this end, we introduce an instance-aware region selection for the training process. Our region selection strategy consists of two stages: instance detection in the first frame, region associations between the first frame and the rest frames.
\begin{figure*}[tbp]
  \centering
  \includegraphics[width=0.875\textwidth]{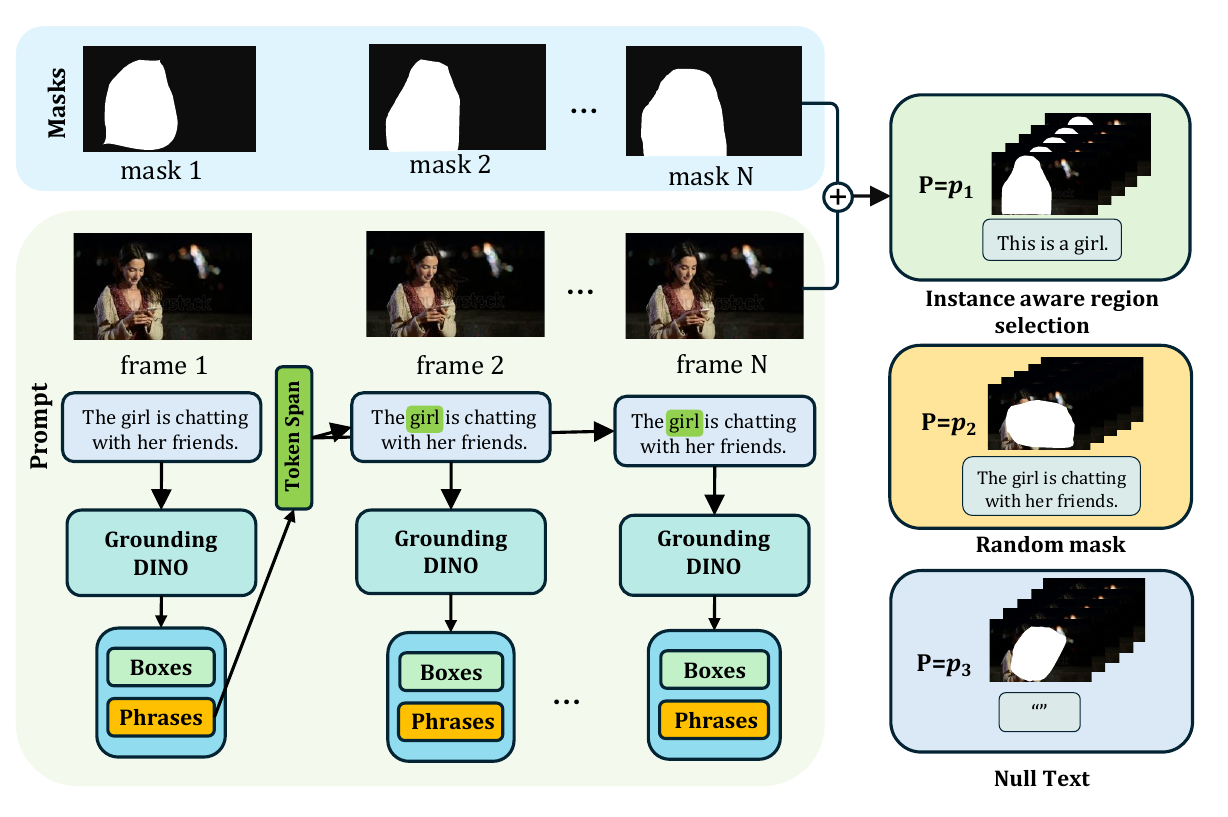} 
  \caption{The instance-aware region selection pipeline and data sampling strategy.  Specifically, we use the \textbf{tokenspan} to fix the candidate phrases and use the random-shaped mask to cover the bounding box. We sample three types of input data with different probabilities when training.}
  \label{fig:data_preprocess}
\end{figure*}
To ground each word or phrase in the text prompt to its corresponding region in the image, we use the GroudningDINO\cite{groundingdino_liu2023grounding} to annotate the first frame in the training data. 
Specifically, we detect first frame and get the returned phrases with bounding boxes. Then we use the \textit{TokenSpan} to force the GroudningDINO detect the bounding box related to the given phrases. This operation guarantee that it will not generate different phrases for a single object. In this way, we can associate the regions in the rest frames with the word or phrases with  the appeared tokens in the first stage. In the training stage, we will randomly sample from three
types of data with different probabilities, including clip with precise masks, clip with random masks and clip with null prompts. 
An illustrated workflow is shown in Figure~\ref{fig:data_preprocess}. \textit{Instance-aware region selection can enable to obtain more precise word and region alignment, and thus achieve better text controllability.} 

\subsection{Adapting Image Generation Model for Video Inpainting}
\label{sec:taskvectorcombination}

The AnimateDiff trains the temporal modules while leaving others parameters frozen. This strategy can have an obvious advantage: the motion module can be easily combined with the different T2Is with the same pretraining base. However, in this way, applying the T2Is directly to our video inpainting model is inconvenient, since the input channels for the video inpainting and the image/video generation model are different. Specifically, the inpainting model takes the latent representation (with 4 channels) of the masked video, mask (with 1 channels) and the latent representation (4 channels) of the noised video clip as the input, while the image/video generation model only requires the latent representation (4 channels) of the noised image as input. The input dimensions between personalized image generation and video inpainting model are mismatched. Moreover, mixing the ability of generation and inpainting is still unexplored before, posing challenge to the availability of the T2I models in the video inpainting tasks.

\begin{figure*}[h]
\vspace{-0.1in}
  \centering
  \includegraphics[width=0.875\textwidth]{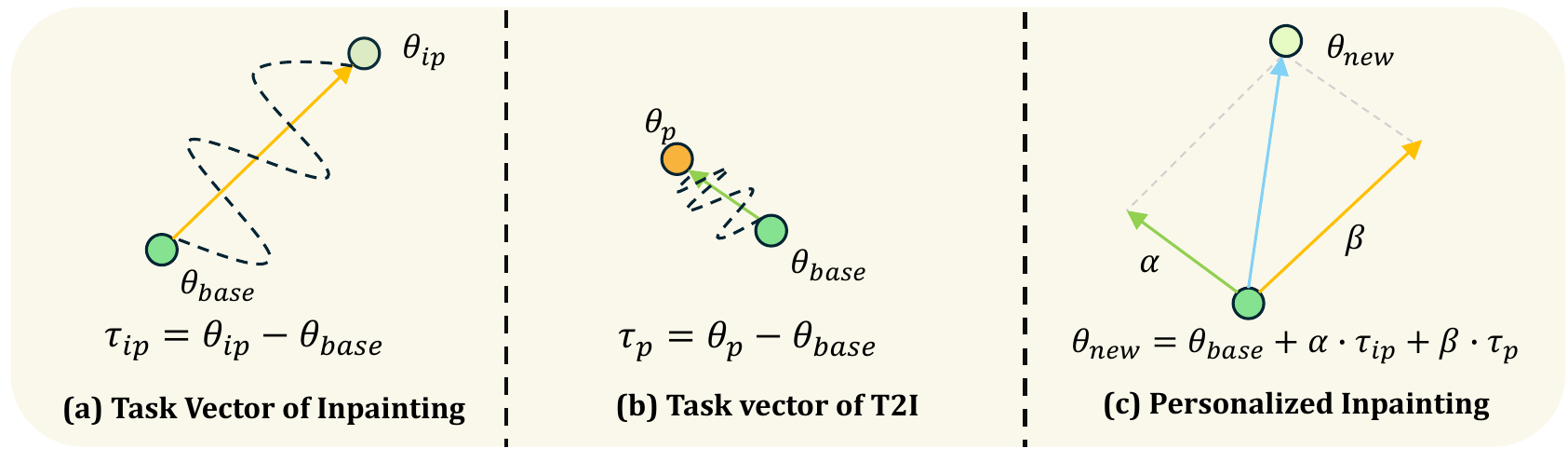} 
  \caption{The pipeline of the our transformation strategy. As shown in the figure, we compute the task vector of inpaintng $\tau_{ip}$ and personalized generation $\tau_{p}$ and subsequently mix the two vectors with the ratio of $\alpha$ and $\beta$ to obtain personalized inpainting model $\theta_{new}$.}
  \label{fig:vector_task}
  \vspace{-0.1in}
\end{figure*}

To solve this problem, we propose a strategy to transform the image generation model into inpainting model inspired by the thought of the \textit{task vector}\cite{taskvector_ilharco2022editing}. First, we pad the input layer of the generation model from 4 channels to 9 channels with zero values to make the two models have the same  dimension. Besides, we donate the padded generation model as the $\theta_{base}$, the padded personalized T2I is $\theta_{p}$ and the inpainting model is $\theta_{ip}$. 
Following the idea of the task vector, we donate the inpainting task vector and the personalized task vector as,
\begin{equation}
    \tau_{ip} = \theta_{ip} - \theta_{base}, \ \ 
    \tau_{p} = \theta_{p} - \theta_{base}
\end{equation}
We add $\tau_{ip}$ and $\tau_{p}$ into the base model and create a new model as,
\begin{equation}
    \theta_{new} = \theta_{base} + \alpha \tau_{ip} + \beta \tau_{p}
\end{equation}
where $\alpha$ and $\beta$ are two hyperparameters. Here, we recommend that $\alpha \in  [0.5,1.5]$, while the $\beta \in [1,2]$. 
The newly created model has the ability to draw personalized visual content in the mask region, while preserving the unmasked region. 

\subsection{Training Objectives}
Given a video clip $v^{1:f}\in \mathbb{R}^{f\times c \times w \times h}$ and its corresponding masked video clip $v_m^{1:f} = v^{1:f}\odot m^{1:f}$, they are encoded to latent codes $z_{0,m}^{f}$ and $z_{0,m}^{1:f}$ frame-wisely by a VAE encoder, where $z_{0,m}^{1:f}, z_0^{1:f} \in \mathbb{R}^{f\times c \times w_1 \times h_1}$. Mask $m$ is resized to $\Bar{m}$, $f$ is the frame number, $c$ is the channel number, $w$ and $h$ are the video width and the video height, while $w_1$ and $w_2$ are width and height of latent codes, respectively.  
In a forward diffusion process, the latent codes $z_0^{1:f}$ is added noise as,
\begin{equation}
    z_t^{1:f} = \sqrt{\Bar{\alpha_t}}\cdot z_0^{1:f} + \sqrt{1-\Bar{\alpha_t}} \cdot \epsilon^{1:f}, \epsilon \sim \mathcal{N}(0,I)
\end{equation}
where $\epsilon^{1:f} \in \mathbb{R}^{f\times c \times w_1 \times h_1}$, $\Bar{\alpha_t}$ is the scalar produced by scheduler in time $t$. 

The UNet model then takes $z_t^{1:f}$, binary mask $\Bar{m}$, masked latents $z_{0,m}$ and the prompt $y$ as the input and predict the added noise $\epsilon$. The final training objective of our model is,
\begin{equation}
    \mathcal{L} = \mathbb{E}_{\mathcal{E}(v^{1:f}), \varepsilon(v_m^{1:f}), \Bar{m}, y, \epsilon^{1:f} \sim \mathcal{N}(0,I),t} [||\epsilon^{1:f} - \epsilon_{\theta}(z_t^{1:f}, \Bar{m}, z_{t,m}^{1:f}, t, c_{\theta}(y))||^{2}_{2}]
\end{equation}
where $\mathcal{E}$ is the VAE encoder and $c_\theta$ is the text encoder, $t$ is the time step, $z_t^{1:f}$ and $z_{t,m}^{1:f}$ are the latent vectors of the video clip and the masked video clip, $\Bar{m}$ is the resized mask with the same shape of $z_t^{1:f}$.
\section{Experiments}
\subsection{Implementation Details}

\textbf{Data.}  We use the WebVid-10M as our training dataset, which contains more than 10M text-video pairs crawled from shutterstock platform. For the data cleaning, we use the \textit{Scenedetect} library to detect the scenes with threshold = 20 and preserve the video which contains only one scene, while discard the video with multiple scenes. To perform instance-aware region selection, we set the detection resolution to $396\times512$ and the threshold for bounding box and phrase selection to $0.2$. We sample the training clip from three types of data as described in Section \ref{sec:instanceawareregionselection} with the probability of 0.7, 0.2 and 0.1, respectively.

\noindent \textbf{Training Details.} We use the AdamW~\cite{adamw_loshchilov2017decoupled} as the optimizer, set the $lr=1e-4$. The type of learning rate scheduler is constant. 
We train the model for 1 epoch. The batch size is 256 by using gradient accumulation. Besides, we use a mixed precision to save GPU memory consumption and accelerate the training process. 
In the training process, we follow DDPM~\cite{ddpm_ho2020denoising}, and use 1000 steps. 
We use the StableDiffusion Inpainting V1.5 as our base model to initialize the spatial block. We train the motion block and leave the spatial block unchanged. For the motion capture module, we initialize the temporal attention layers with AnimateDiff v2, while initialize the damped global attention layer and textual cross-attention layer with Kaiming initialization.  
The training resolution is set to 256$\times$384, sample stride is 4 and the number of frames is 16.

\noindent \textbf{Inference Details.} 
In the inference stage, we follow  DDIM~\cite{ddim_song2020denoising}, use a 50 sampling steps and the classifier-free guidance scale is 14. The mask for per frame can be obtained by Grounding DINO~\cite{groundingdino_liu2023grounding}, Cotracker2~\cite{cotracker_cotracker2karaev2023cotracker}, Xmem~\cite{xmem_cheng2022xmem},
 SAM~\cite{sam_kirillov2023segment} automatically or provided by the users with any shape. The inpainting process can be finished within 1 minute on a Nvidia 3090 GPU.

\subsection{Experimental Results}
We conduct extensive experiments to evaluate our method. In our experiments, we random select 1000 videos from the validation set of the WebVid-10M~\cite{webvid_bain2021frozen}, and extract the first 16 frames in each video with the sample rate of 4. We randomly generate the mask and prompt, and ask model to generate the visual content in the masked region. For baselines, we choose  AnimateDiffV3~\cite{animatediff_guo2023animatediff} and VideoCrater2~\cite{videocrafter2_chen2024videocrafter2} and use the zero-shot inpainting method to fill in the masked region.  Besides, we compare our method with the text-guided inpainting module in VideoComposer~\cite{wang2023videocomposer}. Since AVID~\cite{avid_zhang2023avid} is not open-source when we submit this manuscript, we cannot compare our method with it. We discuss the differences between our model and it in detail in appendix. 
\begin{figure}[t]
  \centering
  \begin{subfigure}{0.19\textwidth}
  \animategraphics[width=1\textwidth]{8}{data/video_org/image_1_}{0}{15}
  \subcaption*{\emph{\scriptsize The yellow dog is }}
  \end{subfigure}
  \begin{subfigure}{0.19\textwidth}
  \animategraphics[width=1\textwidth]{8}{data/vc_images/image_1_}{0}{15}
  \subcaption*{\emph{\scriptsize shaking its head ...}}
  \end{subfigure}
  \begin{subfigure}{0.19\textwidth}
  \animategraphics[width=1\textwidth]{8}{data/vcr_images/image_1_}{0}{15}
  \subcaption*{\emph{\scriptsize }}
  \end{subfigure}
  \begin{subfigure}{0.19\textwidth}
  \animategraphics[width=1\textwidth]{8}{data/sample-1/3286080537ef455be4a91cc0ab56eb61dYIpeJldszIsSfQN-}{0}{15} 
  \subcaption*{\emph{\scriptsize }} 
  \end{subfigure}
  \begin{subfigure}{0.19\textwidth}
   \animategraphics[width=1\textwidth]{8}{data/ours_images/image_1_}{0}{15} 
   \subcaption*{\emph{\scriptsize }} 
   \end{subfigure}
     \begin{subfigure}{0.19\textwidth}
  \animategraphics[width=1\textwidth]{8}{data/video_org_2/ff72c20ccc474ba6d0fc4d182986f021PAntNapq9JUj3KY9-}{0}{15}
  \subcaption*{\emph{\scriptsize The\ \  little\ \  girl\ \  is\  }} 
  \end{subfigure}
  \begin{subfigure}{0.19\textwidth}
  \animategraphics[width=1\textwidth]{8}{data/vc_rank_2_/e6f3328ed43246edacf2cbbce00c0272DfiibCjLmvUIKTmA-}{0}{15}
  \subcaption*{\emph{\scriptsize dressing up ...\ \ \ \ \ \ \ }}
  \end{subfigure}
  \begin{subfigure}{0.19\textwidth}
  \animategraphics[width=1\textwidth]{8}{data/video-crafter-video_2/73340624246b4f279070bf993d1e98d4NOa9P59i4IndHjIP-}{0}{15}
  \subcaption*{\emph{\scriptsize }}
  \end{subfigure}
  \begin{subfigure}{0.19\textwidth}
  \animategraphics[width=1\textwidth]{8}{data/sample-2/371695daf0034955fa9a9b06c50ac040WX5WbaX5b6ngxYgR-}{0}{15}
  \subcaption*{\emph{\scriptsize }}
  \end{subfigure}
  \begin{subfigure}{0.19\textwidth}
   \animategraphics[width=1\textwidth]{8}{data/ours_video_2/ec503596b79042329881c6fa569a315fkewVQ2R0AXp1MmyW-}{0}{15}
   \subcaption*{\emph{\scriptsize }}
   \end{subfigure}
  \begin{subfigure}{0.19\textwidth}
  \animategraphics[width=1\textwidth]{8}{data/video_org_3/f2682ec3-7429-433f-bf69-1f980f121f5f-}{0}{15}
  \subcaption*{\emph{\scriptsize The black car.}}
  
  \caption{Original}
  \end{subfigure}
  \begin{subfigure}{0.19\textwidth}
  \animategraphics[width=1\textwidth]{8}{data/video_composer_rank_3_/e1742387-94ec-4679-b5ef-5869e41866f6-}{0}{15}
  \subcaption*{\emph{\scriptsize }}
  
  \caption{VC}
  \end{subfigure}
  \begin{subfigure}{0.19\textwidth}
  \animategraphics[width=1\textwidth]{8}{data/video_crafter_video_3/e2543fc8cbde4619bcf64984c8af7d91vAQqN1DD1EZpOsM9-}{0}{15}
  \subcaption*{\emph{\scriptsize }}
  \caption{V-Crafter2*}
  \end{subfigure}
  \begin{subfigure}{0.19\textwidth}
  \animategraphics[width=1\textwidth]{8}{data/adiffv3-sample-3/324fa57a-06e5-4d0d-a0ac-cbd387de69cf-}{0}{15}
  \subcaption*{\emph{\scriptsize }}
  
  \caption{AD-V3*}
  \end{subfigure}
  \begin{subfigure}{0.19\textwidth}
   \animategraphics[width=1\textwidth]{8}{data/ours_video_3/c543fd6f-ceaa-4f78-84ba-7b743877d79d-}{0}{15}
   \subcaption*{\emph{\scriptsize }}
   
   \caption{Ours}
   \end{subfigure}
  \caption{The visual comparison results of our \CoCoCo, VideoComposer (VC), Video-Crafter2 (V-Crafter2) and AnimateDiffV3 (AD-V3). $^{*}$ indicates zero-shot inpainting. \emph{Best viewed with Acrobat Reader. Click the images to play the animation clips.}}
  \vspace{-0.175in}
\label{fig:baseline_comparison}
\end{figure} 
\subsubsection{Quantitative  Comparison.}
We use the CLIP score (CS) to measure the ability of text-alignment in different methods. Besides, we use L1 distance to measure background preservation (BP), the lower value means higher preservation in the outside region. Moreover, the cosine similarity is chosen to assess the similarity between the consecutive frames in feature space extracted by CLIP, which is used to measure motion smoothness. 

As shown in Table~\ref{tab:comparison}, our model achieves best over four methods in BP and TC. The BP value of our model is 6.20 in the  scale range [0,255], which is significantly lower than VideoComposer, and outperforms the AnimateDiffV3 and VideoCrafter2. Moreover, our method performs best on temporal consistency , which indicates our model generates more plausible inpainted videos.

For the CLIP-Score, we can find that two Text-to-Video models, AnimateDiffV3 and VideoCrafter2, perform better on the CLIP Score,  which means they can generate the corresponding visual content in the mask region according to the prompt. However, VideoComposer obtains much lower CLIP score than the two zero-shot methods. This is due to it is required to keep the consistency between the generated content and outside region, thus has a weaker ability to keep the textual alignment. Fortunately, our method achieves 24.9 over CLIP score, it is significant higher than VideoComposer, and is close to the AnimateDiffV3. It verifies the effectiveness of our introduced instance-aware region selection strategy and textual cross attention  in the motion capture block.

\begin{table}[t]
    \centering
    \begin{tabular}{ccccccccccc}
        \hline
        & & & \multicolumn{3}{c}{Quantative Results} & &  \multicolumn{4}{c}{User Study} \\
        Method & & &  CS ($\uparrow$) & BP ($\downarrow$) & TC ($\uparrow$)  & & VQ ($\uparrow$) & TA ($\uparrow$) & TC ($\uparrow$) & BP ($\uparrow$) \\
        \hline
        AnimateDiffV3$^{*}$~\cite{animatediff_guo2023animatediff} & & &  25.3 & 7.2 & 96.7 & & 14.6  & 20.8&12.5 & 25.0\\
        VideoCrafter2$^{*}$~\cite{videocrafter2_chen2024videocrafter2} & & & \textbf{26.2} & 7.8 & 96.8 & &2.1 &8.3 &2.1 &6.2 \\
        \hline
        VideoComposer~\cite{wang2023videocomposer} & & & 19.8 &  21.7 & 96.5 & & 12.5 &16.7 &29.2 &4.2 \\ 
        \CoCoCo (ours) & & & 24.9 & \textbf{6.2} & \textbf{97.2} & &\textbf{72.9} &\textbf{54.2} &\textbf{56.2} &\textbf{64.6} \\
        \hline
    \end{tabular}
    \caption{The comparison between our CoCoCo and three other methods. ``CS'', ``BP'',  ``TC'', ``VQ'' and ``TA'' denotes  CLIP score, background preservation, temporal consistency, visual quality, text alignment, respectively. The best results are marked in \textbf{bold}.} 
    \label{tab:comparison}
    \vspace{-24pt}
\end{table}

\subsubsection{Qualitative Results.}
\begin{figure}[htbp]
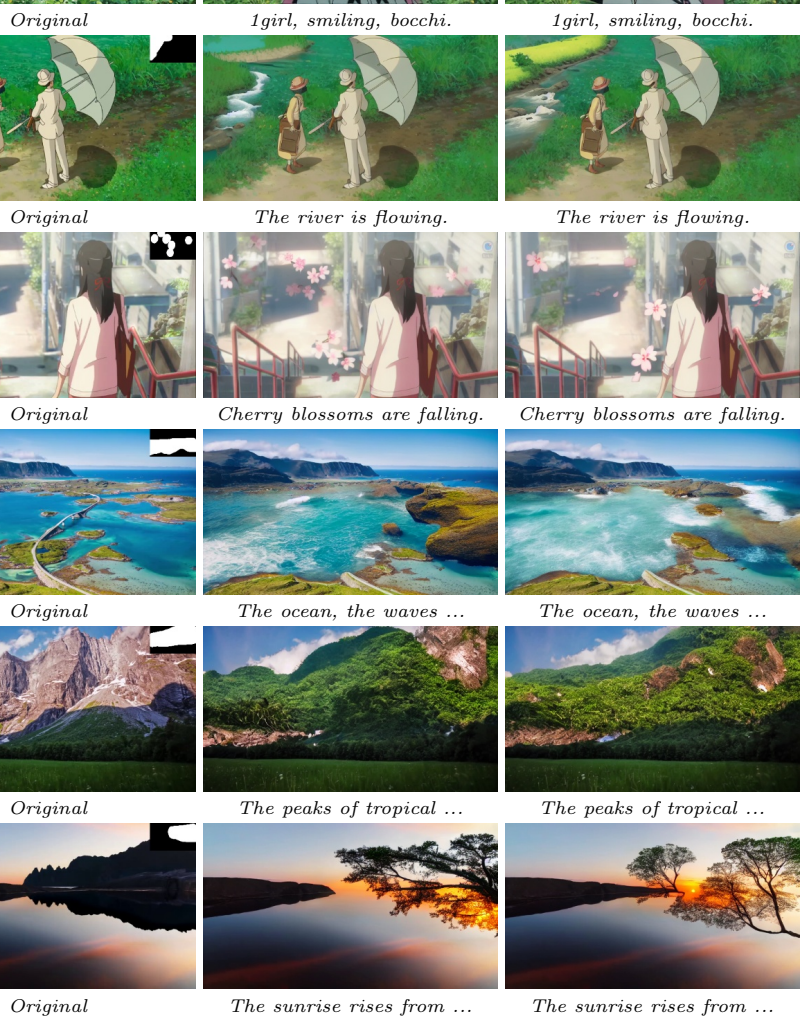

  \centering
\begin{subfigure}{0.32\textwidth}
      \animategraphics[width=1\textwidth]{14}{data/video_3/image_}{0}{14}
      \subcaption*{\emph{\scriptsize Original}}
\end{subfigure}
\begin{subfigure}{0.32\textwidth}
      \animategraphics[width=1\textwidth]{14}{data/video_bocchi_image/video_bocchi_image_1_}{0}{14}
      \subcaption*{\emph{\scriptsize 1girl, smiling, bocchi.}}
\end{subfigure}
\begin{subfigure}{0.32\textwidth}
      \animategraphics[width=1\textwidth]{12}{data/video_bocchi_image/video_bocchi_image_2_}{0}{14}
      \subcaption*{\emph{\scriptsize 1girl, smiling, bocchi.}}
\end{subfigure}

\begin{subfigure}{0.32\textwidth}
      \animategraphics[width=1\textwidth]{12}{data/video_4/image_}{0}{14}
      \subcaption*{\emph{\scriptsize Original}}
\end{subfigure}
\begin{subfigure}{0.32\textwidth}
      \animategraphics[width=1\textwidth]{12}{data/cartoon_river/video_cartoon_river_image_1_}{0}{14}
      \subcaption*{\emph{\scriptsize The river is flowing.}}
\end{subfigure}
\begin{subfigure}{0.32\textwidth}
      \animategraphics[width=1\textwidth]{12}{data/cartoon_river/video_cartoon_river_image_2_}{0}{14}
      \subcaption*{\emph{\scriptsize The river is flowing.}}
\end{subfigure}

\begin{subfigure}{0.32\textwidth}
      \animategraphics[width=1\textwidth]{12}{data/video_5/image_}{0}{14}
      \subcaption*{\emph{\scriptsize Original}}
\end{subfigure}
\begin{subfigure}{0.32\textwidth}
      \animategraphics[width=1\textwidth]{6}{data/cheerry_bloosoms_sanye/video_cartoon_cherry_blossoms_image_1_}{0}{8}
      \subcaption*{\emph{\scriptsize Cherry blossoms are falling.}}
\end{subfigure}
\begin{subfigure}{0.32\textwidth}
      \animategraphics[width=1\textwidth]{6}{data/cheerry_bloosoms_sanye/video_cartoon_cherry_blossoms_image_2_}{0}{8}
      \subcaption*{\emph{\scriptsize Cherry blossoms are falling.}}
\end{subfigure}
\begin{subfigure}{0.32\textwidth}
      \animategraphics[width=1\textwidth]{12}{data/video_6/image_}{0}{15}
      \subcaption*{\emph{\scriptsize Original}}
\end{subfigure}
\begin{subfigure}{0.32\textwidth}
      \animategraphics[width=1\textwidth]{12}{data/sea/video_sea_image_1_}{0}{15}
      \subcaption*{\emph{\scriptsize The ocean, the waves ...}}
\end{subfigure}
\begin{subfigure}{0.32\textwidth}
      \animategraphics[width=1\textwidth]{12}{data/sea/video_sea_image_2_}{0}{15}
      \subcaption*{\emph{\scriptsize The ocean, the waves ...}}
\end{subfigure}
\begin{subfigure}{0.32\textwidth}
      \animategraphics[width=1\textwidth]{8}{data/video_7/image_}{0}{10}
      \subcaption*{\emph{\scriptsize Original}}
\end{subfigure}
\begin{subfigure}{0.32\textwidth}
      \animategraphics[width=1\textwidth]{8}{data/forest/video_forest_image_1_}{0}{10}
      \subcaption*{\emph{\scriptsize The peaks of tropical ...}}
\end{subfigure}
\begin{subfigure}{0.32\textwidth}
      \animategraphics[width=1\textwidth]{8}{data/forest/video_forest_image_2_}{0}{10}
      \subcaption*{\emph{\scriptsize The peaks of tropical ...}}
\end{subfigure}
\begin{subfigure}{0.32\textwidth}
      \animategraphics[width=1\textwidth]{12}{data/video_8/image_}{0}{15}
      \subcaption*{\emph{\scriptsize Original}}
\end{subfigure}
\begin{subfigure}{0.32\textwidth}
      \animategraphics[width=1\textwidth]{12}{data/sunrise/video_sunrise_image_1_}{0}{15}
      \subcaption*{\emph{\scriptsize The sunrise rises from ...}}
\end{subfigure}
\begin{subfigure}{0.32\textwidth}
      \animategraphics[width=1\textwidth]{12}{data/sunrise/video_sunrise_image_2_}{0}{15}
      \subcaption*{\emph{\scriptsize The sunrise rises from ...}}
\end{subfigure}
\caption[]{The inpainting results of our \cococo. The first three rows show results of our model  plugged with customized CounterfeitV3.0 checkpoint or Bocchi LoRA. The last three rows rows are the results without any customized models. \emph{Best viewed with Acrobat Reader. Click the images to play the animation clips.} }
\label{fig:gif_example2}
\end{figure}

We compare our method with baselines by asking users to conduct blind evaluation of different methods on four aspects, including the visual quality (VC), text alignment (TA), temporal consistency (TC) and background preservation (BP). As shown in Table~\ref{tab:comparison}, the videos inpainted by our model are most selected for the best visual quality. Similarly, our method also performs best in the rest of evaluation metrics, especially on the temporal consistency and background preservation. The inpainting results of three baselines and our method are shown in Figure~\ref{fig:baseline_comparison}, VideoComposer can achieve consistency between the masked region and the outside, while the VideoCrafter2 and AnimateDiffV3 do not. However, the background of painted by VideoComposer changes a lot compared with original videos. Our results, as shown in the fifth column, achieve better motion consistency and higher visual quality compared with the other three methods.
More qualitative results of our model can be found in Figure~\ref{fig:gif_example1} and Figure~\ref{fig:gif_example2}.
\begin{table}[t]
    \centering
    \begin{tabular}{cccccc}
        \hline
        Method & & &  CS ($\uparrow$) & BP ($\downarrow$) & TC ($\uparrow$) \\ \hline
        w/o MCB \& IRS & & & 21.6 & 8.3 & 96.3 \\
        w/o MCB & & & 23.6 & 8.6 & 96.8 \\
        \hline
        \CoCoCo (ours) & & & 24.9 &  6.2 & 97.2 \\
        \hline
    \end{tabular}
    \caption{The ablation study results of our model. ``MCB'' and ``IRS'' denote motion capture block and instance-aware region selection. ``CS'', ``BP'',  ``TC'' denotes CLIP score, background preservation, temporal consistency, respectively.}
    \label{tab:ablation_study_comparison}
    \vspace{-26pt}
\end{table}

\begin{figure}[H]
\centering
\begin{subfigure}{0.22\textwidth}
      \animategraphics[width=1\textwidth]{8}{data/ablation_study_org_video/video_org_1_}{0}{14}
      \subcaption*{\emph{\ \ A\ \ yellow\ \ dog\ \ is\ \ }}
\end{subfigure}
\begin{subfigure}{0.22\textwidth}
      \animategraphics[width=1\textwidth]{8}{data/random_mask/random_mask_video_1_}{0}{14}
      \subcaption*{\emph{twisting\ \ its\ \ head...}}
\end{subfigure}
\begin{subfigure}{0.22\textwidth}
      \animategraphics[width=1\textwidth]{8}{data/precise_mask/precise_mask_video_1_}{0}{14}
      \subcaption*{\emph{\scriptsize }}
\end{subfigure}
\begin{subfigure}{0.22\textwidth}
      \animategraphics[width=1\textwidth]{8}{data/ablation_study/video_1_}{0}{14}
      \subcaption*{\emph{\scriptsize }}
\end{subfigure}
\begin{subfigure}{0.22\textwidth}
      \animategraphics[width=1\textwidth]{8}{data/ablation_study_org_video/video_org_2_}{0}{14}
      \subcaption*{\emph{ A green tree...}}
      \subcaption{Original}
\end{subfigure}
\begin{subfigure}{0.22\textwidth}
      \animategraphics[width=1\textwidth]{8}{data/random_mask/random_mask_video_2_}{0}{14}
      \subcaption*{}
      \subcaption{w/o MCB \& IRS}
      \label{fig:ablation_study-a}
\end{subfigure}
\begin{subfigure}{0.22\textwidth}
      \animategraphics[width=1\textwidth]{8}{data/precise_mask/precise_mask_video_2_}{0}{14}
      \subcaption*{}
      \subcaption{w/o MCB}
      \label{fig:ablation_study-b}
\end{subfigure}
\begin{subfigure}{0.22\textwidth}
      \animategraphics[width=1\textwidth]{8}{data/ablation_study/video_2_}{0}{14}
      \subcaption*{}
      \subcaption{Ours}
\end{subfigure}
\caption{The ablation study results of our \cococo. ``MCB'' represents  motion capture block, and ``IRS'' denotes instance-aware region selection. \emph{Best viewed with Acrobat Reader. Click the images to play the animation clips.}}
\label{fig:ablation_study}
\vspace{-24pt}
\end{figure}

\subsubsection{Ablation Study.} 
We conduct ablation study to verify the effectiveness of each component. We have two settings.
For the first setting, we use the motion block with only two temporal attentions but without the damped global attention and the textual cross attention. Meanwhile, we only use random mask selection. For the second setting, we keep the motion block same as the first setting, but replaces the random mask with instance-aware mask region selection. As shown in Figure~\ref{fig:ablation_study}, the model trained under the first setting is not controllable by the given prompts. Contrast to setting 1, setting 2 with the instance-aware mask selection shows higher text-alignment ability.
However,  the second setting without using instance-aware region selection lacks motion consistency between the painted region and outside, and is also opt to generate static object in the inpainting region. This phenomenon indicates that the damped global attention not only captures global motion information but also improves the consistency and visual quality. We also conduct some quantitative experiments to compare the above two settings with our \cococo. The results can be found in Table~\ref{tab:ablation_study_comparison}. We can see from Table~\ref{tab:ablation_study_comparison} that using the instance-aware mask selection can significantly increase CLIP score from 21.6 to 23.6, and adding motion caption block can largely improve  background preservation from 8.6 to 6.2. Using both skills can obviously enhance the temporal consistency from 96.3 to 97.2.

\section{Conclusion}
In this paper, we presented \cococo, a novel text-guided video inpainting model via improving its motion consistency, textual controllability and compatibility with some existing personalized T2Is. To improve motion consistency, we introduced a new damped global attention. Moreover, to achieve better textual controllability, we designed an instance-aware mask region selection module and add an additional textual cross-attention layer to the motion capture block. To leverage existing personalized T2Is, we introduced a task vector combination strategy. Quantitative results and user study experiments showed our method achieved better results compared to its counterparts. More importantly, qualitative results also demonstrated that the proposed model achieved \textit{better motion consistency, textual controllability and model compatibility.}

\clearpage
\bibliographystyle{splncs04}
\bibliography{main}

\clearpage

\appendix

\section{Differences between CoCoCo and AVID}

\begin{table}[h]
\footnotesize
\centering
\begin{tabular}{|c|c|c|c|}
\hline
& \textbf{} & \textbf{AVID} & \textbf{CoCoCo} \\ \hline
\multirow{2}{*}{Data processing}&Video clip & Random & Keep in one scene \\ \cline{2-4}
&Training mask & random & Instance-aware + random \\ \hline
 & \multirow{3}{*}{Attn types} & \multirow{3}{*}{$2\times$Temp attn}  & $2\times$Temp attn \\
& & & $1\times$ DG attn \\
Motion capture & & & $1\times$ Cross attn \\ \cline{2-4}
module& \multirow{2}{*}{Text encoder}  & \multirow{2}{*}{None} & \multirow{2}{*}{CLIP text encoder} \\ 
& & & \\ \hline
Transformation & & \multirow{2}{*}{\ding{55}} & \multirow{2}{*}{\checkmark} \\
strategy & & & \\ \hline
Sparse control & & \checkmark & \checkmark \\ \hline
\end{tabular}
\caption{The comparison between CoCoCo and AVID. ``DG Attn'' and ``Temp Attn'' denote the Damped Global Attention and Temporal Attention, respectively.}
\label{tab:comparison2}
\end{table}

In this discussion, we highlight the distinctions between our CoCoCo and AVID\cite{avid_zhang2023avid}. Initially, as demonstrated in Table \ref{tab:comparison2}, AVID adopts a random approach for selecting video clips and generating training masks. Conversely, CoCoCo not only selects video clips from a single scene but also employs a blend of precise and random masks. This method significantly diminishes training noise and enhances the consistency of the visual content produced by CoCoCo. Additionally, we have revamped the motion capture module. Unlike AnimateDiff\cite{animatediff_guo2023animatediff} and AVID, which utilize only two temporal attentions, our system incorporates four attention layers: two temporal attentions, one damped global attention, and one cross attention. Crucially, AVID's motion module lacks the capability to incorporate text information. We address this limitation by integrating an additional cross attention block, enabling direct infusion of text input into the motion module. 
Lastly, AVID does not study the T2I\cite{huggingface,civitai} transformation strategy and thus it could not use the personalized T2I models. 
We introduce a straightforward but effective strategy to merge personalized T2Is with an inpainting model, assessing the new model's effectiveness within our CoCoCo framework. 

\section{Understanding of T2I Transformation Strategy}

\subsection{The similarity between the parameters in Generation and Inpainting Model}

\begin{figure*}[h]
  \centering
  \includegraphics[width=0.895\textwidth]{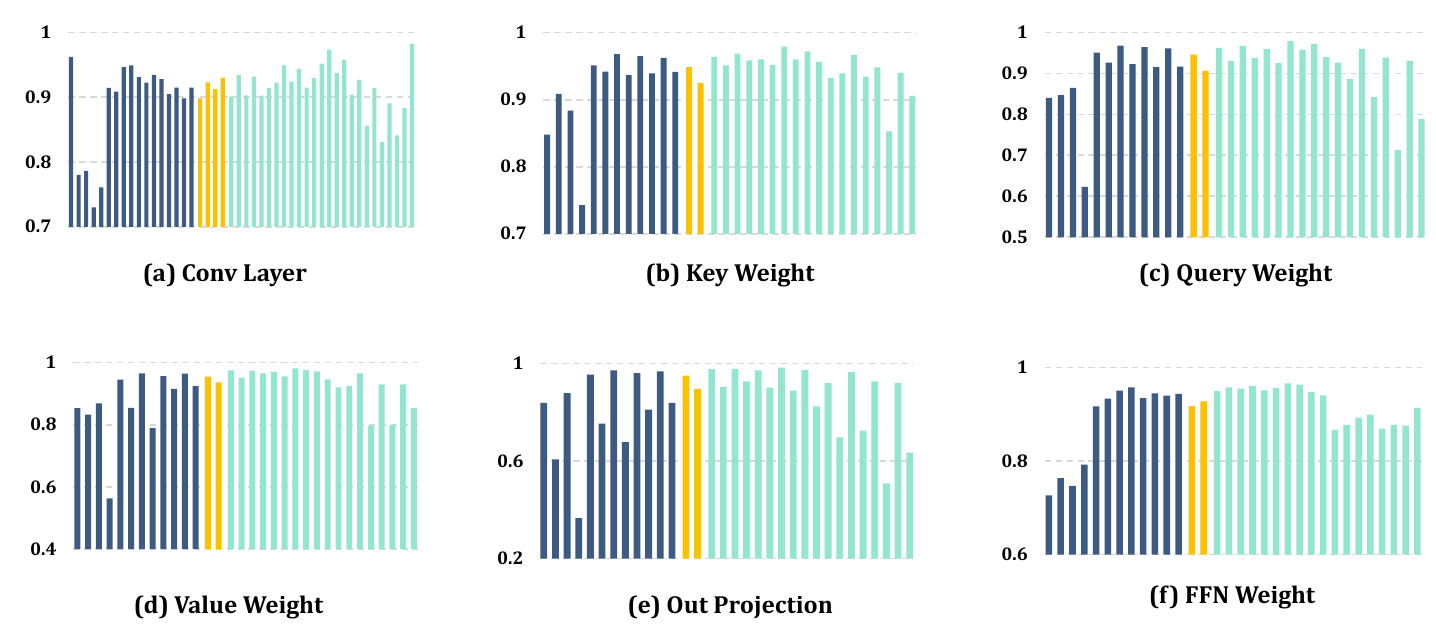}
  \caption{The inpainting model's modifications relative to the generation model span six different layer types: convolutional (conv), query, key, value, output projection, and feedforward network (ffn) layers. A higher value indicates lower difference between the matrices in the inpainting and generation models. The blue bar signifies cosine similarity in the down block, the orange bar denotes cosine similarity in the middle block, and the green bar represents cosine similarity between two matrices in the up block. Note that the up blocks have more layers than down blocks.} 
  \label{model_difference}
\end{figure*} 

As illustrated in Figure \ref{model_difference}, our analysis reveals a significant resemblance between most parameters in the image generation and inpainting models. Specifically, the largest differences occur in the shallow and output layers of the two models, while the remaining layers exhibit over 0.95 similarity. This minimal modification allows the inpainting model to achieve inpainting capabilities while maintaining generalization to personalized Text-to-Image (T2I) models. This examination explains how two models, despite differing in parameters, training objectives, and even architecture, can still produce a valid task vector and remain compatible with the personalized image generation model.

\subsection{The parameter sensitivity of T2I Transformation Strategy}

\begin{figure}
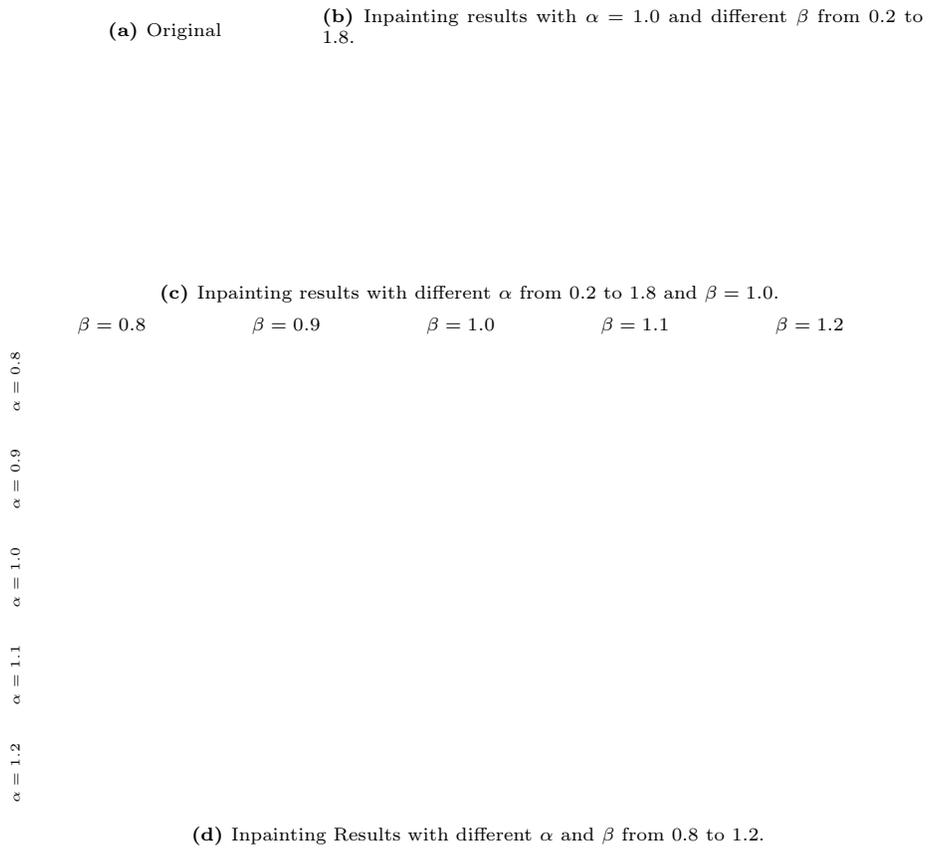

\centering

\begin{minipage}{0.325\textwidth}
\begin{subfigure}{1.0\textwidth}
    \animategraphics[width=1\textwidth]{12}{data/video_4/image_}{0}{15}
    \subcaption{Original}
\end{subfigure}
\end{minipage}
\begin{minipage}{0.655\textwidth}
\begin{subfigure}{0.235\textwidth}
    \animategraphics[width=1\textwidth]{12}{data/video_param_sensitivity_alpha_0.2_beta_1.0_image/video_param_sensitivity_alpha_0.2_beta_1.0_image_}{0}{15}
\end{subfigure}
\begin{subfigure}{0.235\textwidth}
    \animategraphics[width=1\textwidth]{12}{data/video_param_sensitivity_alpha_0.6_beta_1.0_image/video_param_sensitivity_alpha_0.6_beta_1.0_image_}{0}{15} 
    \end{subfigure}
\begin{subfigure}{0.235\textwidth}
    \animategraphics[width=1\textwidth]{12}{data/video_param_sensitivity_alpha_0.8_beta_1.0_image/video_param_sensitivity_alpha_0.8_beta_1.0_image_}{0}{15} 
    \end{subfigure}
\begin{subfigure}{0.235\textwidth}
    \animategraphics[width=1\textwidth]{12}{data/video_param_sensitivity_alpha_1.0_beta_1.0_image/video_param_sensitivity_alpha_1.0_beta_1.0_image_}{0}{15} 
    \end{subfigure}

\begin{subfigure}{0.235\textwidth}
    \animategraphics[width=1\textwidth]{12}{data/video_param_sensitivity_alpha_1.2_beta_1.0_image/video_param_sensitivity_alpha_1.2_beta_1.0_image_}{0}{15} 
    \end{subfigure}
\begin{subfigure}{0.235\textwidth}
    \animategraphics[width=1\textwidth]{12}{data/video_param_sensitivity_alpha_1.4_beta_1.0_image/video_param_sensitivity_alpha_1.4_beta_1.0_image_}{0}{15} 
    \end{subfigure}
\begin{subfigure}{0.235\textwidth}
    \animategraphics[width=1\textwidth]{12}{data/video_param_sensitivity_alpha_1.8_beta_1.0_image/video_param_sensitivity_alpha_1.8_beta_1.0_image_}{0}{15} 
    \end{subfigure}
    \subcaption{Inpainting results with $\alpha=1.0$ and different $\beta$ from 0.2 to 1.8.}
\end{minipage}

\begin{minipage}[h]{0.975\textwidth} 
\begin{subfigure}{0.22\textwidth} 
      \animategraphics[width=1\textwidth]{12}{data/video_param_sensitivity_alpha_1.0_beta_0.2_image/video_param_sensitivity_alpha_1.0_beta_0.2_image_}{0}{15}
    \end{subfigure}
\begin{subfigure}{0.22\textwidth} 
      \animategraphics[width=1\textwidth]{12}{data/video_param_sensitivity_alpha_1.0_beta_0.6_image/video_param_sensitivity_alpha_1.0_beta_0.6_image_}{0}{15}
    \end{subfigure}
\begin{subfigure}{0.22\textwidth} 
      \animategraphics[width=1\textwidth]{12}{data/video_param_sensitivity_alpha_1.0_beta_0.8_image/video_param_sensitivity_alpha_1.0_beta_0.8_image_}{0}{15}
    \end{subfigure}
\begin{subfigure}{0.22\textwidth} 
      \animategraphics[width=1\textwidth]{12}{data/video_param_sensitivity_alpha_1.0_beta_1.0_image/video_param_sensitivity_alpha_1.0_beta_1.0_image_}{0}{15}
    \end{subfigure}
    
\begin{subfigure}{0.22\textwidth} 
      \animategraphics[width=1\textwidth]{12}{data/video_param_sensitivity_alpha_1.0_beta_1.2_image/video_param_sensitivity_alpha_1.0_beta_1.2_image_}{0}{15}
    \end{subfigure}
\begin{subfigure}{0.22\textwidth} 
      \animategraphics[width=1\textwidth]{12}{data/video_param_sensitivity_alpha_1.0_beta_1.4_image/video_param_sensitivity_alpha_1.0_beta_1.4_image_}{0}{15}
    \end{subfigure}
\begin{subfigure}{0.22\textwidth} 
      \animategraphics[width=1\textwidth]{12}{data/video_param_sensitivity_alpha_1.0_beta_1.8_image/video_param_sensitivity_alpha_1.0_beta_1.8_image_}{0}{15}
    \end{subfigure}
    \subcaption{Inpainting results with different $\alpha$ from 0.2 to 1.8 and $\beta=1.0$.}
\end{minipage}

\begin{minipage}{0.01\textwidth}
\rotatebox{90}{\tiny $\alpha=1.2$\ \ \ \ \ \ \tiny $\alpha=1.1$\ \ \ \ \ \ \tiny $\alpha=1.0$\ \ \ \ \ \ \tiny $\alpha=0.9$\ \ \ \ \ \ $\alpha=0.8$}
\end{minipage}
\begin{minipage}{0.98\textwidth}
\begin{subfigure}{0.185\textwidth}
    \subcaption*{$\beta=0.8$}
      \animategraphics[width=1\textwidth]{12}{data/video_param_sensitivity_alpha_0.8_beta_0.8_image/video_param_sensitivity_alpha_0.8_beta_0.8_image_}{0}{15} 
    \end{subfigure}
    \begin{subfigure}{0.185\textwidth}
      \subcaption*{$\beta=0.9$}
      \animategraphics[width=1\textwidth]{12}{data/video_param_sensitivity_alpha_0.9_beta_0.8_image/video_param_sensitivity_alpha_0.9_beta_0.8_image_}{0}{15} 
    \end{subfigure}
    \begin{subfigure}{0.185\textwidth}
      \subcaption*{$\beta=1.0$}
      \animategraphics[width=1\textwidth]{12}{data/video_param_sensitivity_alpha_1.0_beta_0.8_image/video_param_sensitivity_alpha_1.0_beta_0.8_image_}{0}{15} 
    \end{subfigure}
    \begin{subfigure}{0.185\textwidth}
      \subcaption*{$\beta=1.1$}
      \animategraphics[width=1\textwidth]{12}{data/video_param_sensitivity_alpha_1.1_beta_0.8_image/video_param_sensitivity_alpha_1.1_beta_0.8_image_}{0}{15} 
    \end{subfigure}
    \begin{subfigure}{0.185\textwidth}
      \subcaption*{$\beta=1.2$}
      \animategraphics[width=1\textwidth]{12}{data/video_param_sensitivity_alpha_1.2_beta_0.8_image/video_param_sensitivity_alpha_1.2_beta_0.8_image_}{0}{15} 
    \end{subfigure}

    \begin{subfigure}{0.185\textwidth}
      \animategraphics[width=1\textwidth]{12}{data/video_param_sensitivity_alpha_0.8_beta_0.9_image/video_param_sensitivity_alpha_0.8_beta_0.9_image_}{0}{15} 
    \end{subfigure}
    \begin{subfigure}{0.185\textwidth}
      \animategraphics[width=1\textwidth]{12}{data/video_param_sensitivity_alpha_0.9_beta_0.9_image/video_param_sensitivity_alpha_0.9_beta_0.9_image_}{0}{15} 
    \end{subfigure}
    \begin{subfigure}{0.185\textwidth}
      \animategraphics[width=1\textwidth]{12}{data/video_param_sensitivity_alpha_1.1_beta_0.9_image/video_param_sensitivity_alpha_1.1_beta_0.9_image_}{0}{15} 
    \end{subfigure}
    \begin{subfigure}{0.185\textwidth} 
      \animategraphics[width=1\textwidth]{12}{data/video_param_sensitivity_alpha_1.1_beta_0.9_image/video_param_sensitivity_alpha_1.1_beta_0.9_image_}{0}{15}
    \end{subfigure}
    \begin{subfigure}{0.185\textwidth} 
      \animategraphics[width=1\textwidth]{12}{data/video_param_sensitivity_alpha_1.2_beta_0.9_image/video_param_sensitivity_alpha_1.2_beta_0.9_image_}{0}{15}
    \end{subfigure}

    \begin{subfigure}{0.185\textwidth}
      \animategraphics[width=1\textwidth]{12}{data/video_param_sensitivity_alpha_0.8_beta_1.0_image/video_param_sensitivity_alpha_0.8_beta_1.0_image_}{0}{15} 
    \end{subfigure}
    \begin{subfigure}{0.185\textwidth}
      \animategraphics[width=1\textwidth]{12}{data/video_param_sensitivity_alpha_0.9_beta_1.0_image/video_param_sensitivity_alpha_0.9_beta_1.0_image_}{0}{15} 
    \end{subfigure}
    \begin{subfigure}{0.185\textwidth}
      \animategraphics[width=1\textwidth]{12}{data/video_param_sensitivity_alpha_1.0_beta_1.0_image/video_param_sensitivity_alpha_1.0_beta_1.0_image_}{0}{15} 
    \end{subfigure}
    \begin{subfigure}{0.185\textwidth}
      \animategraphics[width=1\textwidth]{12}{data/video_param_sensitivity_alpha_1.1_beta_1.0_image/video_param_sensitivity_alpha_1.1_beta_1.0_image_}{0}{15} 
    \end{subfigure}
    \begin{subfigure}{0.185\textwidth}
      \animategraphics[width=1\textwidth]{12}{data/video_param_sensitivity_alpha_1.2_beta_1.0_image/video_param_sensitivity_alpha_1.2_beta_1.0_image_}{0}{15} 
    \end{subfigure}

    \begin{subfigure}{0.185\textwidth}
      \animategraphics[width=1\textwidth]{12}{data/video_param_sensitivity_alpha_0.8_beta_1.1_image/video_param_sensitivity_alpha_0.8_beta_1.1_image_}{0}{15} 
    \end{subfigure}
    \begin{subfigure}{0.185\textwidth}
      \animategraphics[width=1\textwidth]{12}{data/video_param_sensitivity_alpha_0.9_beta_1.1_image/video_param_sensitivity_alpha_0.9_beta_1.1_image_}{0}{15} 
    \end{subfigure}
    \begin{subfigure}{0.185\textwidth}
      \animategraphics[width=1\textwidth]{12}{data/video_param_sensitivity_alpha_1.0_beta_1.1_image/video_param_sensitivity_alpha_1.0_beta_1.1_image_}{0}{15} 
    \end{subfigure}
    \begin{subfigure}{0.185\textwidth}
      \animategraphics[width=1\textwidth]{12}{data/video_param_sensitivity_alpha_1.1_beta_1.1_image/video_param_sensitivity_alpha_1.1_beta_1.1_image_}{0}{15} 
    \end{subfigure}
    \begin{subfigure}{0.185\textwidth}
      \animategraphics[width=1\textwidth]{12}{data/video_param_sensitivity_alpha_1.1_beta_1.2_image/video_param_sensitivity_alpha_1.1_beta_1.2_image_}{0}{15} 
    \end{subfigure}

    \begin{subfigure}{0.185\textwidth}
      \animategraphics[width=1\textwidth]{12}{data/video_param_sensitivity_alpha_0.8_beta_1.2_image/video_param_sensitivity_alpha_0.8_beta_1.2_image_}{0}{15} 
    \end{subfigure}
    \begin{subfigure}{0.185\textwidth}
      \animategraphics[width=1\textwidth]{12}{data/video_param_sensitivity_alpha_0.9_beta_1.2_image/video_param_sensitivity_alpha_0.9_beta_1.2_image_}{0}{15} 
    \end{subfigure}
    \begin{subfigure}{0.185\textwidth}
      \animategraphics[width=1\textwidth]{12}{data/video_param_sensitivity_alpha_1.0_beta_1.2_image/video_param_sensitivity_alpha_1.0_beta_1.2_image_}{0}{15} 
    \end{subfigure}
    \begin{subfigure}{0.185\textwidth}
      \animategraphics[width=1\textwidth]{12}{data/video_param_sensitivity_alpha_1.1_beta_1.2_image/video_param_sensitivity_alpha_1.1_beta_1.2_image_}{0}{15} 
    \end{subfigure}
    \begin{subfigure}{0.185\textwidth}
      \animategraphics[width=1\textwidth]{12}{data/video_param_sensitivity_alpha_1.2_beta_1.2_image/video_param_sensitivity_alpha_1.2_beta_1.2_image_}{0}{15} 
    \end{subfigure}
    \subcaption{Inpainting Results with different $\alpha$ and $\beta$ from 0.8 to 1.2.}
\end{minipage}

    \caption{The parameter sensitivity of transformation strategy in our CoCoCo. \emph{Best viewed with Acrobat Reader. Click the images to play the animation clips.} }
    \label{fig:param_sensitivity}
\end{figure}

In this section, we explore the parameter sensitivity for our transformation strategy with the T2I model (Counterfeit V30) and the Inpainting model (SD Inpainting 1.5\cite{stablediffusion_blattmann2023align}). It's important to note that the optimal values for \(\alpha\) and \(\beta\) may vary when utilizing different T2I and Inpainting models. As depicted in Figure \ref{fig:param_sensitivity}, a lower \(\alpha\) value slightly alters the background of video clips, yet the visual content within the masked area remains of high quality. Conversely, as \(\beta\) increases, the inpainted style begins to more closely resemble the style of the specified checkpoint. Crucially, setting \(\alpha\) and \(\beta\) to high values can cause the model to malfunction. In contrast, a lower \(\beta\) value results in a model that functions akin to a standard inpainting model. Meanwhile, setting \(\alpha=0\) transforms the model into an image generation model.

\section{More Examples}

In this section, we give more examples that demonstrate uncropping (outpainting), retexturing, and swapping using a precise mask or random mask. These examples of uncrop specifically employ the sparse control model from AnimateDiff to guide the generation of details. Although the sparse control model was initially developed for AnimateDiff V3, our experiments illustrate its compatibility with our CoCoCo framework, showcasing its versatility and effectiveness in enhancing visual content through detailed inpainting.

\begin{figure}
    \centering
    \begin{minipage}{0.02\textwidth}
      \rotatebox{90}{\scriptsize \ \ UnCrop}
    \end{minipage}
    \begin{minipage}{0.96\textwidth}
    \begin{subfigure}{0.3\textwidth}
      \animategraphics[width=1\textwidth]{12}{data/uncrop_org/image_}{0}{15}
      \subcaption*{Prompt: Sky and river.}
    \end{subfigure}
    \begin{subfigure}{0.3\textwidth}
      \animategraphics[width=1\textwidth]{12}{data/uncrop_video_uncrop_image/video_uncrop_image_1_}{0}{15}
      \subcaption*{}
    \end{subfigure}
    \begin{subfigure}{0.3\textwidth}
      \animategraphics[width=1\textwidth]{12}{data/uncrop_video_uncrop_image/video_uncrop_image_2_}{0}{15}
      \subcaption*{}
    \end{subfigure}
    \end{minipage}

    \begin{minipage}{0.02\textwidth}
      \rotatebox{90}{\scriptsize \ \ \ Retexturing }
    \end{minipage}
    \begin{minipage}{0.96\textwidth}
    \begin{subfigure}{0.3\textwidth}
      \animategraphics[width=1\textwidth]{12}{data/retexturing/image_}{0}{15}
      \subcaption*{Prompt: A blue car.}
    \end{subfigure}
    \begin{subfigure}{0.3\textwidth}
      \animategraphics[width=1\textwidth]{12}{data/video_car_image/video_car_image_1_}{0}{15}
      \subcaption*{}
    \end{subfigure}
    \begin{subfigure}{0.3\textwidth}
      \animategraphics[width=1\textwidth]{12}{data/video_car_image/video_car_image_2_}{0}{15}
      \subcaption*{}
    \end{subfigure}
    \end{minipage}

    \begin{minipage}{0.02\textwidth}
      \rotatebox{90}{\scriptsize \ \ \ \ Swapping\ \ \ \ }
    \end{minipage}
    \begin{minipage}{0.96\textwidth}
    \begin{subfigure}{0.3\textwidth}
      \animategraphics[width=1\textwidth]{12}{data/video_10/image_}{0}{15}
      \subcaption*{Prompt: A meadow.}
    \end{subfigure}
    \begin{subfigure}{0.3\textwidth}
      \animategraphics[width=1\textwidth]{12}{data/meadow6/video_meadow_image_1_}{0}{15}
      \subcaption*{}
    \end{subfigure}
    \begin{subfigure}{0.3\textwidth}
      \animategraphics[width=1\textwidth]{12}{data/meadow6/video_meadow_image_2_}{0}{15}
      \subcaption*{}
    \end{subfigure}

    \begin{subfigure}{0.3\textwidth}
      \animategraphics[width=1\textwidth]{12}{data/video_11/image_}{0}{15}
      \subcaption*{Prompt: Cherry blossoms.}
    \end{subfigure}
    \begin{subfigure}{0.3\textwidth}
      \animategraphics[width=1\textwidth]{12}{data/cherry_blossoms/video_cherry_blossoms_image_1_}{0}{15}
      \subcaption*{}
    \end{subfigure}
    \begin{subfigure}{0.3\textwidth}
      \animategraphics[width=1\textwidth]{12}{data/cherry_blossoms/video_cherry_blossoms_image_2_}{0}{15}
      \subcaption*{}
    \end{subfigure}

    \begin{subfigure}{0.3\textwidth}
      \animategraphics[width=1\textwidth]{12}{data/video_12/image_}{0}{15}
      \subcaption*{Prompt: The river with ice.}
    \end{subfigure}
    \begin{subfigure}{0.3\textwidth}
      \animategraphics[width=1\textwidth]{12}{data/river/video_river_image_1_}{0}{15}
      \subcaption*{}
    \end{subfigure}
    \begin{subfigure}{0.3\textwidth}
      \animategraphics[width=1\textwidth]{12}{data/river/video_river_image_2_}{0}{15}
      \subcaption*{}
    \end{subfigure}
    \end{minipage}

    \caption{More Experimental Results. The first row is the examples of uncropping. The second row is the examples of retexturing. The third row to fifth row are examples for swapping. \emph{Best viewed with Acrobat Reader. Click the images to play the animation clips.}}
    
    \label{fig:enter-label}
\end{figure}

\end{document}